\documentclass[nohyperref]{article}

\usepackage{microtype}
\usepackage{graphicx}
\usepackage{subfigure}
\usepackage{booktabs} 
\usepackage{enumitem}
\usepackage{hyperref}
\usepackage{svg}
\usepackage{float}

\usepackage[accepted]{icml2023}

\usepackage{amsmath}
\usepackage{amssymb}
\usepackage{mathtools}
\usepackage{amsthm}
\usepackage{verbatim}

\usepackage[capitalize,noabbrev]{cleveref}

\newcommand{\blackboard}[0]{\emph{DTM}}

\newcommand{\car}{\texttt{car}}
\newcommand{\cdr}{\texttt{cdr}}
\newcommand{\cons}{\texttt{cons}}

\theoremstyle{plain}

\theoremstyle{definition}

\theoremstyle{remark}

\usepackage[textsize=tiny]{todonotes}

\icmltitlerunning{Differentiable Tree Operations Promote Compositional Generalization}

\begin{document}

\twocolumn[
\icmltitle{Differentiable Tree Operations Promote Compositional Generalization}



\icmlsetsymbol{equal}{*}
\icmlsetsymbol{msri}{$\dagger$}

\begin{icmlauthorlist}
\icmlauthor{Paul Soulos}{jhu,msri}
\icmlauthor{Edward Hu}{mila,msri}
\icmlauthor{Kate McCurdy}{edin,msri}
\icmlauthor{Yunmo Chen}{jhu-cs,msri}
\icmlauthor{Roland Fernandez}{msr}
\icmlauthor{Paul Smolensky}{jhu,msr}
\icmlauthor{Jianfeng Gao}{msr}
\end{icmlauthorlist}

\icmlaffiliation{jhu}{Department of Cognitive Science, Johns Hopkins University, Baltimore, MD, USA}
\icmlaffiliation{mila}{Mila, Université de Montreal, Montreal, CA}
\icmlaffiliation{edin}{School of Informatics, University of Edinburgh, Edinburgh, UK}
\icmlaffiliation{jhu-cs}{Department of Computer Science, Johns Hopkins University, Baltimore, MD, USA}
\icmlaffiliation{msr}{Microsoft Research, Redmond, WA, USA}

\icmlcorrespondingauthor{Paul Soulos}{psoulos1@jhu.edu}

\icmlkeywords{Compositionality, Generalization, Neurosymbolic, Natural Language Processing, Differentiable Computing, Trees, Vector Symbolic Architecture, Hyperdimensional Computing}

\vskip 0.3in
]



\printAffiliationsAndNotice{}  

\begin{abstract}
In the context of structure-to-structure transformation tasks, learning sequences of discrete symbolic operations poses significant challenges due to their non-differentiability. To facilitate the learning of these symbolic sequences, we introduce a differentiable tree interpreter that compiles high-level symbolic tree operations into subsymbolic matrix operations on tensors. We present a novel Differentiable Tree Machine (\blackboard) architecture that integrates our interpreter with an external memory and an agent that learns to sequentially select tree operations to execute the target transformation in an end-to-end manner. With respect to out-of-distribution compositional generalization on synthetic semantic parsing and language generation tasks, \blackboard\ achieves 100\% while existing baselines such as Transformer, Tree Transformer, LSTM, and Tree2Tree LSTM achieve less than 30\%. \blackboard\ remains highly interpretable in addition to its perfect performance.
\end{abstract}

\section{Introduction}
\begin{figure}[ht]
\vskip 0.2in
\begin{center}
\centerline{\includegraphics[width=1.0\columnwidth]{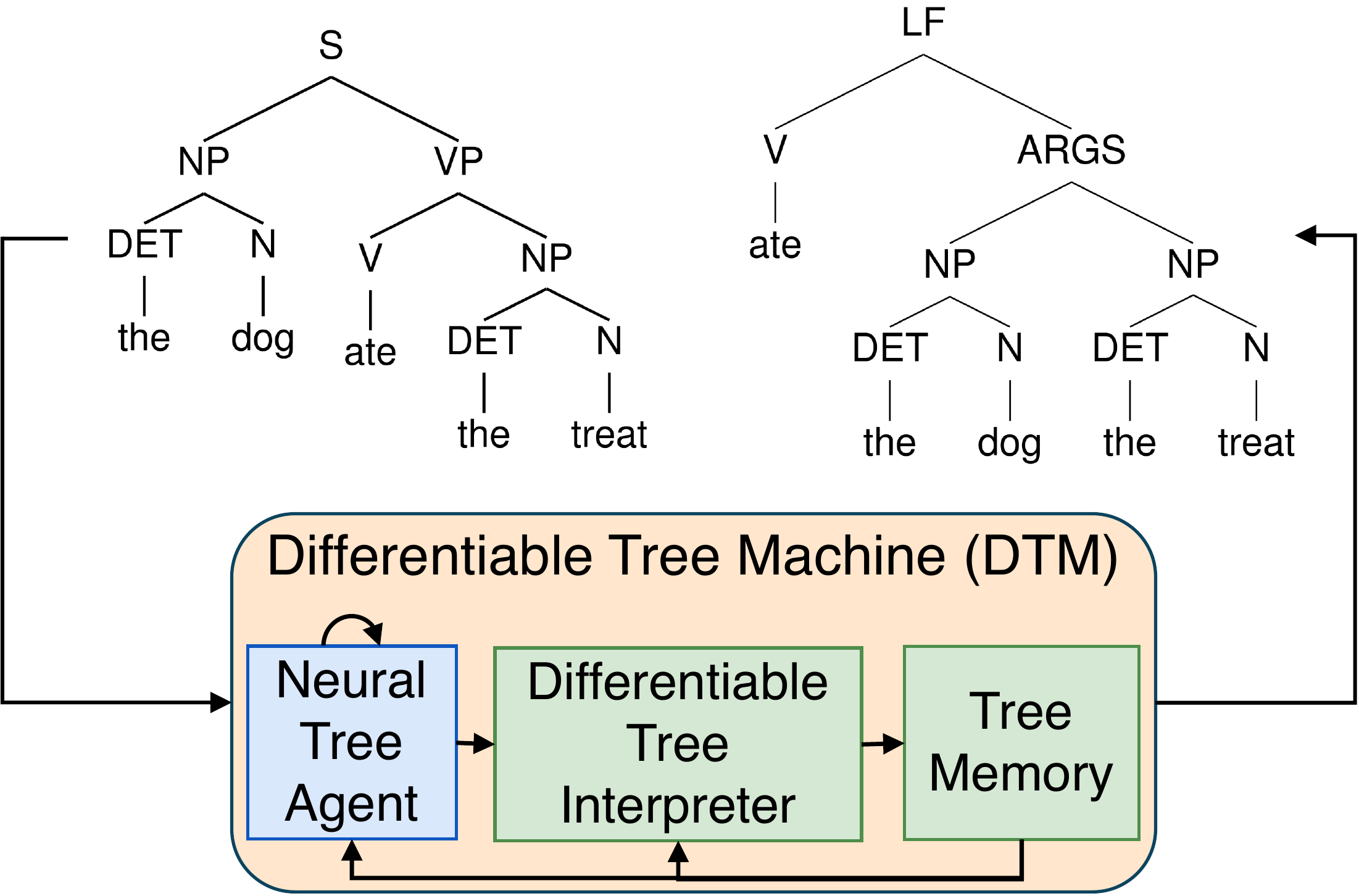}}
\caption{A high level overview of our model which consists of three modules. The Neural Tree Agent is a learnable component which, at each step of processing, selects the operation to perform and the arguments over which to operate. The Differentiable Tree Interpreter is a closed-form function precomputed at initialization which compiles high level symbolic operations into subsymbolic matrix operations on tensors. The output of the interpreter is a blended tree that is written to Tree Memory, which functions as a working memory to hold various partial and candidate trees. The final tree written to memory is the output tree. Blue represents the component with learnable parameters, and green represents components that operate in tree space. }
\label{fig:high-level}
\end{center}
\vskip -0.2in
\end{figure}

Symbolic manipulation is a hallmark of human reasoning \cite{Newell1980PhysicalSS, newell1982knowledge}. Reasoning within the symbolic space through discrete symbolic operations can lead to improved out-of-distribution (OOD) generalization and enhanced interpretability. Despite the significant advances in representation learning made by modern deep learning, learning to directly manipulate discrete symbolic structures remains a challenge. One key issue is the non-differentiability of discrete symbolic operations, which makes them incompatible with gradient-based learning methods. Continuous representations offer greater learning capacity, but often at the expense of interpretability and compositional generalization.

Tensor Product Representation (TPR) provides a general encoding of structured symbolic objects in vector space \cite{Smolensky1990TensorPV}. TPR decomposes a symbolic object into a set of role-filler pairs, such as a position in a tree (the role) and the label for that position (the filler of that role). The role and filler in each pair are represented by vectors and bound together using the tensor product operation.

\begin{figure}[t]
\vskip 0.2in
\begin{center}
\centerline{\includegraphics[width=\columnwidth]{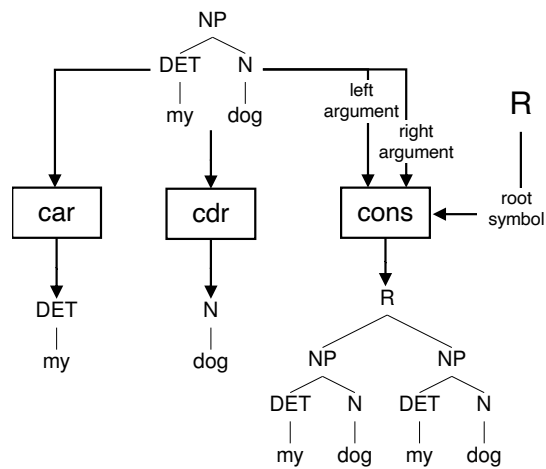}}
\caption{An example showing the output of our three operations.}
\label{fig:car-cdr-cons}
\end{center}
\vskip -.2in
\end{figure}

In this work, we focus on binary trees and three Lisp operators: \car, \cdr, and \cons\ \cite{steele1990common} (also known as left-child, right-child, and construct new tree). Examples of these operations are shown in Figure \ref{fig:car-cdr-cons}. Crucially, within the TPR space, these symbolic operators on discrete objects become linear operators on continuous vectors (\textsection \ref{sec:tpr}). Unlike normal symbolic structures, the vector space nature of TPRs allows blending multiple symbolic structures as interpolations between classic discrete structures. We restrict processing over our TPR encodings to the interpretable linear operations implementing the three Lisp operators and their interpolations, making the computation differentiable and accessible to backpropagation. Gradients can flow through our differentiable tree operations, allowing us to optimize the sequencing and blending of linear operations using nonlinear deep learning models to parameterize the decision space.

Employing TPRs to represent binary trees, we design a novel Differentiable Tree Machine architecture, \blackboard\footnote{Code available at \url{https://github.com/psoulos/dtm}.} (\textsection \ref{sec:model}), capable of systematically manipulating binary trees (overview shown in Figure \ref{fig:high-level}). At each step of computation, \blackboard\ soft-selects a binary tree to read from an external memory, soft-selects a linear operator to apply to the tree, and then writes the resulting tree to a new memory slot. Soft-selecting among a set $S$ of $n$ elements in a vector space entails computing a vector $w \in \mathbb{R}^n$ of non-negative weights that sum to one and returning the sum of the elements in $S$ weighted by $w$ (i.e., $w \cdot S$). As learning progresses, our experiments show that, without explicit pressure to do so, the weights on the operators tend to become 1-hot, and the continuous blends of trees become more discrete as we converge to a discrete sequence of operations for each sample. We validate our proposal empirically on a series of synthetic tree-to-tree datasets that test a model's ability to generalize compositionally (\textsection \ref{sec:empirical}).

The \blackboard\ architecture achieves near-perfect out-of-distribution generalization for the examined synthetic tree-transduction tasks, on which previous models such as Transformers, LSTMs, and their tree variants exhibit weak or no out-of-distribution generalization. 

In summary, our contributions include:

\begin{itemize}
  \item A novel \blackboard\ architecture for interpretable, continuous manipulation of blended binary trees.
  \item A dataset with four tasks to test out-of-distribution generalization on binary tree-to-tree tasks.
  \item Empirical comparison of \blackboard\ and baselines on these datasets which demonstrates the unique advantages of \blackboard\ in terms of compositional generalization and interpretability.
  \item Ablation experiments showing how different aspects of \blackboard\ contribute to its success.
\end{itemize}

\section{Related Work}

\subsection{Compositional Generalization}
Research on compositional generalization has been one of the core issues in Machine Learning since its inception. Despite improvements in architectures and scalability \cite{csordas2021devil}, neural network models still struggle with out-of-distribution generalization \cite{najoung}. The lack of robust compositional generalization has been a central argument against neural networks as models of cognition for almost half a century by proponents of GOFAI systems that leverage symbolic structures \citep[e.g.,][]{fodor1988connectionism,marcus2003algebraic}. These symbolic systems are brittle and face scalability problems because their nondifferentiability makes them incompatible with gradient learning methods. Our work attempts to bridge the neural network-symbolic divide by situating symbolic systems in vector space, where a first-order gradient can be derived as a learning signal. 

In practice, the term ``compositional generalization" has been associated with a range of different tasks \citep{hupkes2020compositionality}. \citet{kim-linzen-2020-cogs} identify a key distinction relevant to natural language: lexical versus structural generalization. \emph{Lexical} generalization is required when a model encounters a primitive (e.g., a word) in a structural environment (e.g., a position in a tree) where it has not been seen during training. \emph{Structural} generalization is required when a model encounters a structure that was not seen during training, such as a longer sentence or a syntactic tree with new nodes. \citet{najoung} demonstrate that structural and lexical generalization remain unsolved: pretrained language models still do not consistently generalize fully to novel lexical items and structural positions. The tasks we study below explicitly test both types of compositional generalization (\textsection \ref{sec:datasets}).

Our proposed \blackboard\ model encodes and manipulates data exclusively in the form of Tensor Product Representations (TPRs; \textsection \ref{sec:TPR}). This formalism inherently supports composition and decomposition through symbol-role bindings, creating an inductive bias toward symbolic operations. \emph{Lexical} generalization is straightforward when syntactic trees are encoded as TPRs: a novel symbol can easily bind to any role. \emph{Structural} generalization is possible through our linear representation of the \car, \cdr, and \cons\ functions, as these operators are not sensitive to the size or structure of the trees they take as arguments.
We evaluate \blackboard's capacity for both types of compositional generalization in \textsection \ref{sec:results}.

\subsection{Tensor Product Representations (TPRs)} \label{sec:TPR}
Tensor Product Representations have been used to enhance performance and interpretability across textual question-answering \cite{schlag2018learning, palangi2018question}, natural-language-to-program-generation \cite{chen}, math problem solving \cite{imanol}, synthetic sequence tasks \cite{mccoy2018rnns, soulos2020discovering}, summarization \cite{jiang2021enriching}, and translation \cite{soulos2021structural}. While previous work has focused on using TPRs to structure and interpret representations, the processing over these representations was done using black-box neural networks. In this work, we predefine structural operations to process TPRs and use black-box neural networks to parameterize the information flow and decision making in our network. 

\subsection{Vector Symbolic Architectures}
Vector Symbolic Architecture (VSA) \cite{gayler2003vsa_jackendoff, plate, kanerva2009hyperdimensional} is a computing framework for realizing classic symbolic algorithms in vector space. Our work bridges VSAs and Deep Learning by using black-box neural networks to write differentiable vector-symbolic programs. For a recent survey on VSAs, see \citet{kleyko2022}, and for VSAs with spiking neurons see \citet{eliasmith}.

\subsection{Differentiable Computing}
One approach to integrating neural computation and GOFAI systems is Differentiable Computing. In this approach, components of symbolic computing are re-derived in a continuous and fully differentiable manner to faciliate learning with backpropagation. In particular, neural networks that utilize an external memory have received considerable attention \cite{Graves2014NeuralTM, Graves2016HybridCU, Weston2014MemoryN, nram}.

Another significant aspect of Differentiable Computing involves integrating structured computation graphs into neural networks. Tree-LSTMs \cite{tai-etal-2015-improved, dong2016language, chen2018tree} use parse trees to encode parent nodes in a tree from their children's representations or decode child nodes from their parent's representations. Some Transformer architectures modify standard multi-headed attention to integrate tree information \cite{wang-etal-2019-tree, sartran2022transformer}, while other Transformer architectures integrate tree information in the positional embeddings \cite{shiv2019novel}. Neural Module Networks \cite{Andreas2015NeuralMN} represent a separate differentiable computing paradigm, where functions in a symbolic program are replaced with black-box neural networks. 

A few works have explored using differentiable interpreters to learn subfunctions from program sketches and datasets \cite{bosnjak17a, Reed2015NeuralP}. Most similar to our work, \citet{Joulin2015InferringAP} and \citet{Grefenstette2015LearningTT} learn RNNs capable of leveraging a stack with discrete push and pop operations in a differentiable manner. While they use a structured object to aid computation, the operations they perform involve read/write operations over unstructured vectors, whereas the operations we deploy in this work consist of structured operations over vectors with embedded structure.

\section{Differentiable Tree Operations} \label{sec:tpr}

A completely general formalization of compositional structure is defined by a set of roles, and an instance of a structure results from assigning these roles to particular fillers \cite{Newell1980PhysicalSS}. Intuitively, a role characterizes a position in the structure, and its filler is the substructure that occupies that position. 

In this work, we use a lossless encoding for structure in vector space. Given a tree depth limit of depth $D$, the total number of  tree nodes is $N=(b^{D+1}-1)/(b-1)$ where $b$ is the branching factor. We generate a set of $N$ orthonormal role vectors of dimension $d_r=N$. For a particular position $r_i$ in a tree, a filler $f_i$ is assigned to this role by taking the outer product of the embedding vectors for the filler and the role: $f_i \otimes r_i$. The embedding of the entire structure is the sum over the individual filler-role combinations $T = \sum_{i=1}^N f_i \otimes r_i$. Since the role vectors are orthonormal, a filler $f_i$ can be recovered from $T$ by the inner product between $T$ and $r_i$, $f_i = Tr_i$.

Moving forward, we will focus on the case of binary trees ($b = 2$), which serve as the foundation for a substantial amount of symbolic AI research. From the orthonormal role set, we can generate matrices to perform the Lisp operators \car, \cdr, and \cons. 
For a tree node reached from the root by following the path $x$, denote its role vector by $r_x$; e.g., $r_{011}$ is the role vector reached by descending from the root to the left (0th) child, then the right (1st) child, then the right (1st) child. Let $P = \{r_x \| \hspace{.3em} \lvert x \rvert < D\} $ be the roles for all paths from the root down to a depth less than $D$.

In order to extract the subtree which is the left child of the root (Lisp \car), we need to zero out the root node and the right child subtree while moving each filler in the left subtree up one level. Extracting the right subtree (Lisp \cdr) is a symmetrical process. This can be accomplished by:

$\car(T)\!=\!D_0 T$;
$\cdr(T)\!=\!D_1 T$;
$D_c \!=\! I_F \! \otimes \! \sum_{x \in P} r_{x}r_{cx}^\top$

where $I$ is the identity matrix on filler space.

Lisp \cons\ constructs a new binary tree given two trees to embed as the left- and right-child. In order to add a subtree as the $c$th child of a new root node, we define $E_c$ to add $c$ to the top of the path-from-the-root for each position:

$\cons(T_0,T_1)=E_0 T_0 + E_1 T_1$;
$E_c = I_F \otimes \sum_{x \in P} r_{cx}r_x^\top$

 When performing \cons, a new filler $s$ can be placed at the parent node of the two subtrees $T_0$ and $T_1$ by adding $s \otimes r_{root}$ to the output of \cons.
Our model uses linear combination to blend the results of applying the three Lisp operations. The output of step $l\in1:L$, when operating on the arguments $\vec{T}^{(l)}=(T_{\car}^{(l)},T_{\cdr}^{(l)},T_{\cons 0}^{(l)},T_{\cons1}^{(l)})$, is\:

\begin{equation} \label{eq:output}
\begin{split}
O^{(l)}(\vec{w}^{(l)}, \vec{T}^{(l)}, s^{(l)}) = w^{(l)}_\car \car(T_{\car}^{(l)}) + w^{(l)}_\cdr \cdr(T_{\cdr}^{(l)}) \\ + w^{(l)}_\cons \left( \cons(T_{\cons0}^{(l)}, T_{\cons1}^{(l)}) +  s^{(l)} \otimes r_{root} \right)
\end{split}
\end{equation}

The three operations are weighted by the level-specific weights $\vec{w}^{(l)} = (w^{(l)}_\car,  w^{(l)}_\car, w^{(l)}_\cons)$, which sum to 1.

\section{Differentiable Tree Machine (\blackboard) Architecture for Binary Tree Transformation} \label{sec:model}

\begin{figure*}[ht]
\vskip 0.2in
\begin{center}
\centerline{\includegraphics[width=\textwidth]{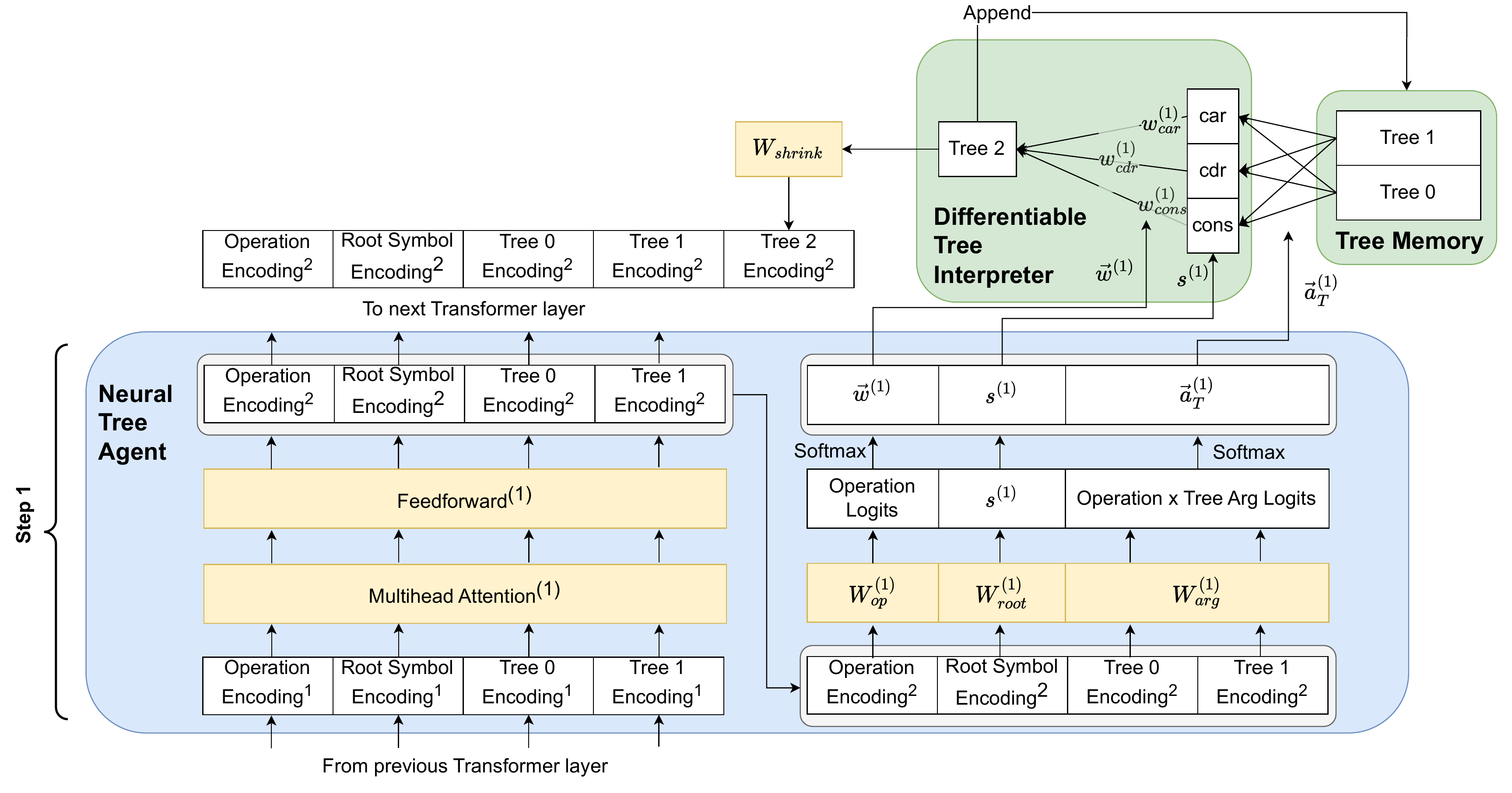}}
\caption{Step 1 of the \blackboard\ architecture is expanded to show the information flow. The yellow boxes identify the parameters that are learnable. The blue box highlights the Neural Tree Agent, and the green boxes highlight components in tree space: the Differentiable Tree Interpreter (Eq 1) and Tree Memory. The left side of the Neural Tree Agent is a standard transformer layer with self-attention and a feedforward network. Residual connections and layer norm are not shown.}
\label{architecture}
\end{center}
\vskip -0.2in
\end{figure*}

\begin{figure}[ht]
\vskip 0.2in
\begin{center}
\centerline{\includegraphics[width=\columnwidth]{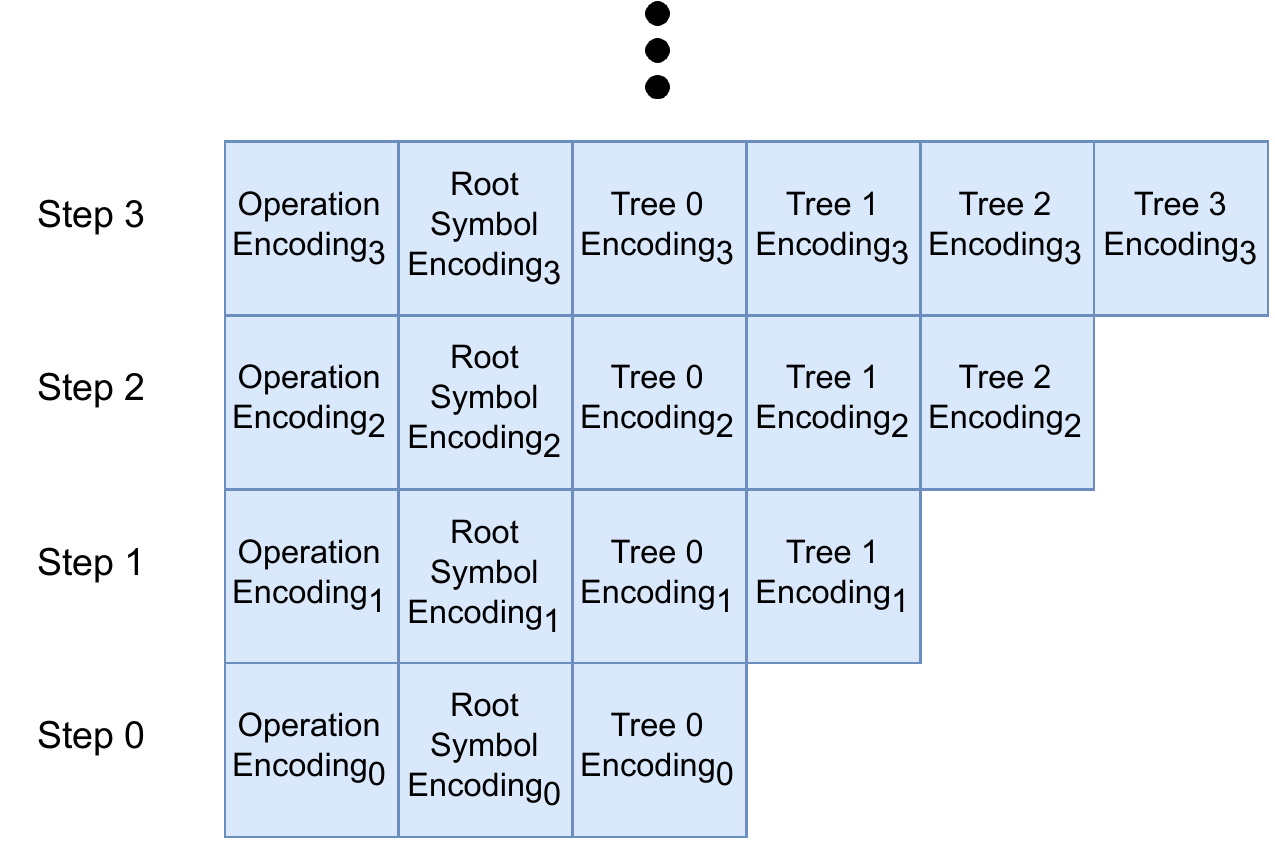}}
\caption{Inputs to the Neural Tree Agent at each step of processing. The length of the input grows by one token each step to incorporate the output of the previous step.}
\label{step}
\end{center}
\vskip -0.2in
\end{figure}

In order to actualize the theory described in Section \ref{sec:tpr}, we introduce the Differentiable Tree Machine (\blackboard), a model that is capable of learning how to perform operations over binary trees. Since the primitive functions \car, \cdr, and \cons\ are precomputed at initialization from the orthogonally generated role vectors, this learning problem reduces to learning which operations to perform on which trees in Tree Memory to arrive at a correct output. A high-level overview of our model is given in Figure \ref{fig:high-level}. \blackboard\ consists of a learned component (Neural Tree Agent), a differentiable pre-designed tree interpreter described in Equation \ref{eq:output}, and an external Tree Memory for storing trees.

At a given timestep $l$, our agent selects the inputs to Equation \ref{eq:output}: the tree arguments for the operations ($\vec{T}^{(l)}$), the new root symbol for \cons\ ($s^{(l)}$) and how much to weight the output of each operation ($\vec{w}^{(l)}$). To select $\vec{T}^{(l)}$, \blackboard\ produces coefficients over the trees in Tree Memory, where the coefficients across trees in $\vec{T}^{(l)}$ sum to 1. For example, if Tree Memory contains only $T_0\ \&\  T_1$, weights $\vec{a}^{(l)}_{\car} = (a^{(l)}_{\car,0}, a^{(l)}_{\car,1})$ are computed to define the argument to \car:
$T_{\car}^{(l)}=a^{(l)}_{\car,0}T_0 + a^{(l)}_{\car,1}T_1$,
and similarly for \cdr\ and the two arguments of \cons.  $\vec{a}^{(l)}_{T}=(\vec{a}^{(l)}_{\car};\vec{a}^{(l)}_{\cdr};\vec{a}^{(l)}_{\cons 0};\vec{a}^{(l)}_{\cons 1})$ denotes all such weights.

These decisions are computed within the Neural Tree Agent module of \blackboard\ using a standard Transformer layer \cite{vaswani} consisting of multiheaded self-attention, a feedforward network, residual connections, and layer norm. Figure \ref{architecture} shows the computation in a single step of \blackboard. When a binary tree is read from Tree Memory, it is compressed from the TPR dimension $d_{tpr}$ to the Transformer input dimension $d_{model}$ using a linear transformation $W_{shrink} \in \mathbb{R}^{d_{tpr} \times d_{model}}$. We also feed in two special tokens to encode the operation-weighting coefficients and the new root-symbol prediction. In addition to the standard parameters in a Transformer layer, our model includes three additional weight matrices $W_{op} \in \mathbb{R}^{d_{model} \times 3}$, $W_{root} \in \mathbb{R}^{d_{model} \times d_{symbol}}$, and $W_{arg} \in \mathbb{R}^{d_{model} \times 4}$. $W_{op}$ projects the operation token encoding into logits for the three operations which are then normalized via softmax. $W_{root}$ projects the root symbol token encoding into the new root symbol. $W_{arg}$ projects the encoding of each TPR in memory to logits for the four tree arguments, the input to \car, \cdr, and \cons\ left and right. The arguments for each operator are a linear combination of all the TPRs in memory, weighted by the softmax of the computed logits. These values are used to create the output for this step as described in equation \ref{eq:output} and the output TPR is written into Tree Memory at the next sequential slot. For the beginning of the next step, the contents of the Tree Memory are encoded to model dimension by $W_{shrink}$ and appended to the Neural Tree Agent Transformer input sequence. The input to the Neural Tree Agent grows by one compressed tree encoding at each time step to incorporate  the newly produced tree, as shown in Figure \ref{step}.

The tree produced by the final step of our network is used as the output (predicted tree). We minimize the mean-squared error between the predicted symbol at each node in the predicted tree and the target tree for all non-empty nodes in the target tree. We penalize the norm of filled nodes in the predicted tree that are empty in the target tree. Additional training details can be found in Section \ref{sec:blackboard-training}.

\section{Empirical Validation} \label{sec:empirical}

\begin{table*}[ht]
\vskip 0.15in
\begin{center}
\begin{sc}
\begin{tabular}{lccccc}
\toprule
Data set & \blackboard\ & Transformer & LSTM & Tree2Tree & Tree Transformer \\
\midrule
\textbf{car\_cdr\_seq}\\
 -train & .95 ± .04 & 1.0 ± .00 & 1.0 ± .00 & 1.0 ± .00 & 1.0 ± .00 \\
 -test IID & .95 ± .04 & 1.0 ± .00 & 1.0 ± .00 & 1.0 ± .00 & 1.0 ± .00  \\
 -test OOD lexical & \textbf{.94 \textpm\ .04} & .66 ± .00 & .66 ± .00 & .66 ± .00 & .66 ± .00 \\
 -test OOD structural & \textbf{.93 ± .04} & .68 ± .01  & .47 ± .03 & .74 ± .02 & .64 ± .01 \\
\textbf{Active$\leftrightarrow$Logical} & & &  \\
 -train &  1.0 ± .00 & 1.0 ± .00 & 1.0 ± .00 & 1.0 ± .00 & 1.0 ± .00  \\
 -test IID &  1.0 ± .00 & 1.0 ± .00 & 1.0 ± .00 & .99 ± .00 & 1.0 ± .00 \\
-test OOD lexical &  \textbf{1.0 ± .00} & .00 ± .00 & .00 ± .00 & .00 ± .00 & .00 ± .00 \\
-test OOD structural &   \textbf{1.0 ± .00} & .00 ± .00 & .00 ± .00  & .10 ± .03 & .03 ± .01 \\
\textbf{Passive$\leftrightarrow$Logical} & & &  \\
 -train & 1.0 ± .00 & 1.0 ± .00 & 1.0 ± .00 & 1.0 ± .00 & 1.0 ± .00 \\
 -test IID &  1.0 ± .00 & 1.0 ± .00 & 1.0 ± .00 & 1.0 ± .00 & 1.0 ± .00 \\
-test OOD lexical &   \textbf{1.0 ± .00} & .00 ± .00 & .00 ± .00 & .00 ± .00 & .00 ± .00 \\
-test OOD structural &   \textbf{1.0 ± .00} & .00 ± .00 & .00 ± .00 & .19 ± .02 & .00 ± .00  \\
\textbf{Active \& Passive$\rightarrow$Logical} & & &  \\
 -train & 1.0 ± .00 & 1.0 ± .00 & 1.0 ± .00 & 1.0 ± .00 & 1.0 ± .00 \\
 -test IID & 1.0 ± .00 & 1.0 ± .00 & 1.0 ± .00 & .99 ± .00 & 1.0 ± .00 \\
 -test OOD lexical &   \textbf{1.0 ± .00} & .00 ± .00 & .00 ± .00 & .00 ± .00 & .00 ± .00 \\
-test OOD structural &   \textbf{1.0 ± .00} & .00 ± .00 & .00 ± .00 & .10 ± .01 & .01 ± .00 \\
\bottomrule
\end{tabular}
\end{sc}
\end{center}
\vskip -0.1in
\caption{Mean accuracy and standard deviation across five random initializations on synthetic tree-to-tree transduction tasks using different model architectures. Test sets include in-distribution and out-of-distribution splits.}
\label{tab:results}
\end{table*}

\subsection{Datasets} \label{sec:datasets}
We introduce the \textbf{Basic Sentence Transforms} dataset for testing tree-to-tree transformations.\footnote{Data available at \url{https://huggingface.co/datasets/rfernand/basic_sentence_transforms}.} It contains various synthetic tree-transform tasks, including a Lisp function interpreter task and several natural-language tasks inspired by semantic parsing and language generation. This dataset is designed to test compositional generalization in structure transformations, as opposed to most existing compositionality-related datasets, which focus on linear sequence transformations.

Each task in the dataset has five splits: train, validation, test, out-of-distribution lexical (OOD-lexical), and out-of-distribution structural (OOD-structural). The OOD-lexical split tests a model's ability to perform zero-shot lexical generalization to new adjectives not seen during training. The OOD-structural split tests a model's structural generalization by using longer adjective sequences and new tree positions not encountered during training. The train split has 10,000 samples, while the other splits have 1,250 samples each. Samples of these tasks are shown in Appendix \ref{sec:dataset-samples} and additional information about the construction of the dataset is in Appendix \ref{sec:dataset-construction}. We focus our evaluation on the following four tasks:

\textbf{CAR-CDR-SEQ} is a tree transformation task where the source tree represents a template-generated English sentence, and the target tree represents a subset of the source tree. The target tree is formed from a sequence of Lisp \car\ and \cdr\ operations on the source tree. The desired sequence of operations is encoded into a single token in the source tree root, and the transformation requires learning how to interpret this root token and execute the associated sequence of actions. Although its internal structure is not accessible to the model, the token is formed according to the Lisp convention for combining these operations into a single symbol (starting with a $\mathtt{c}$, followed by the second letter of each operation, and terminated by an $\mathtt{r}$, e.g., $\mathtt{cdaadr}$ denotes the operation sequence: \cdr, \car, \car, \cdr). This task uses sequences of 1-4 \car/\cdr\ operations (resulting in 30 unique functions).

\textbf{Active$\leftrightarrow$Logical} contains syntax trees in active voice and logical form. Transforming from active voice into logical form is similar to semantic parsing, and transducing from logical form to active voice is common in natural language generation. An example from this task is shown in Figure \ref{fig:high-level}.

\textbf{Passive$\leftrightarrow$Logical}
contains syntax trees in passive voice and logical form. This task is similar to the one above but is more difficult and requires more operations. The passive form also contains words that are not present in logical form, so unlike Active$\leftrightarrow$Logical, the network needs to insert additional nodes. At first glance, this does not seem possible with \car, \cdr, and \cons; we will show how our network manages to solve this problem in an interpretable manner in \textsection \ref{sec:interpret}. An example from this task is shown in Figure \ref{fig:interpret}.

\textbf{Active \& Passive$\rightarrow$Logical} contains input trees in either active or passive voice and output trees in logical form. This tests whether a model can learn to simultaneously parse different types of trees, distinguished  by their structures, into a shared logical form. 

\subsection{Baseline Architectures}
We compare against standard seq2seq \cite{sutskever} models and tree2tree models as our baselines. For seq2seq models, we linearize our trees by coding them as left-to-right sequences with parentheses to mark the tree structure. Our seq2seq models are an Encoder-Decoder Transformer \cite{vaswani} and an LSTM \cite{Hochreiter1997LongSM}. We test two tree2tree models: Tree2Tree LSTM \cite{chen2018tree} and Tree Transformer \citet{shiv2019novel}. Tree2Tree LSTM combines a Tree-LSTM encoder \cite{tai-etal-2015-improved} and a Tree-LSTM decoder \cite{dong2016language}. Tree Transformer encodes tree information in relative positional embeddings as the path from one node to another. Training details for the baselines can be found in Appendix A.

\subsection{Results} \label{sec:results}
The results for \blackboard\ and the baselines can be seen in Table \ref{tab:results}. \blackboard\ achieves 100\% accuracy across all splits for three of the four tasks, and for some of the runs in the  \textbf{CAR\_CDR\_SEQ} task. While the baselines perform similarly to \blackboard\ when compared on train and test IID, the results are drastically different when comparing the results across OOD splits. Across all tasks, \blackboard\ generalizes similarly regardless of the split, whereas the baselines struggle with lexical generalization and fail completely at structural generalization.

The baseline models perform the best on \textbf{CAR\_CDR\_SEQ}, whereas this is the most difficult task for \blackboard. We suspect that tuning the hyperparameters for \blackboard\ directly on this task would alleviate the less-than-perfect performance. Despite performing less than perfectly, \blackboard\ performance on the OOD splits of \textbf{CAR\_CDR\_SEQ} outperforms all of the baselines. Whereas \textbf{CAR\_CDR\_SEQ} involves identifying a subtree (or subsequence for the baselines) within the input, the other four tasks involve reorganizing the input and potentially adding additional tokens in the case of \textbf{Passive$\leftrightarrow$Logical}. On these linguistically-motivated tasks, the baselines mostly achieve 0\% OOD generalization, with a maximum of 19\%.

\blackboard\ can be compared against the other tree models to see the effects of structured processing in vector space. While the Tree2Tree LSTM and Tree Transformer are both capable of representing trees, the processing that occurs over these trees is still black-box nonlinear transformations. \blackboard\ isolates black-box nonlinear transformations to the Neural Tree Agent, while the processing over trees is factorized into interpretable operations over tree structures with excellent OOD generalization. This suggests that it is not the tree encoding scheme itself that is critical, but rather the processing that occurs over the trees.

\subsection{Ablations} \label{sec:ablations}
In order to examine how the components of our model come together to achieve compositional generalization, we run several ablation experiments on \textbf{Active$\leftrightarrow$Logical}.

\subsubsection{Pre-defined structural operations}
What purpose do the \car, \cdr, and \cons\ equations defined in Section \ref{sec:tpr} play in our network's success?
Instead of defining the transformations with the equations, we can randomly initialize the $D$ and $E$ matrices and allow them to be learned during training. The results of learning the $D$ and $E$ matrices are shown in Table \ref{tab:predefined}. Since the $D$ and $E$ matrices, whether predefined or learned, operate only on the role space, it is unsurprising that our model continues to achieve perfect lexical generalization without the predefined equations for $D$ and $E$. However, structural generalization suffers dramatically when the $D$ and $E$ matrices are learned. This result indicates that the predefined tree operations are essential for our model to achieve structural generalization.

\begin{table}[t]
\begin{center}
\begin{sc}
\resizebox{\columnwidth}{!}{
\begin{tabular}{lp{8em}p{8em}}
\toprule
& Predefined \mbox{transformations} &  Learned \mbox{transformations} \\
\midrule
-train & $1.0 \pm .00$ & $1.0 \pm .00$ \\
-test IID & $1.0 \pm .00$ &  $.99 \pm .02$ \\
-lexical & $1.0 \pm .00$ & $.99 \pm .01$ \\
-Structural & $1.0 \pm .00$ & $.35 \pm .08$ \\
\bottomrule
\end{tabular}
}
\end{sc}
\end{center}
\vskip -0.1in
\caption{Accuracy on Active$\leftrightarrow$Logical across five random initializations for models with predefined \car, \cdr\ and \cons\ operations versus learned transformations. Lexical and structural are test OOD splits.}
\label{tab:predefined}
\end{table}

\subsubsection{Blending vs.\ discrete selections}
While our model converges to one-hot solutions where it chooses a single operation over a single tree in memory, it is not constrained to do so, and it deploys heavy blending prior to final convergence. There are two sources of blending: the input arguments to each operation can be a blend of trees in memory, and the output written to memory is a weighted blend of the three operations. We can explore the importance of blending by restricting our model to make discrete decisions using the Gumbel-Softmax distribution \cite{jang2017categorical,maddison2017the}. Table \ref{tab:gumbel} shows the results of models trained with (blend-producing) softmax or (discrete) Gumbel-Softmax for argument and operation selection. We observe that the use of Gumbel-Softmax in either operation or argument sampling leads to a complete breakdown in performance. This demonstrates that blending is an essential component of our model, and that our network is not capable of learning the task without it.

\begin{table*}[h]
\begin{center}
\begin{sc}
\begin{tabular}{lp{7.5em}p{7.5em}p{7.5em}p{7.5em}p{0em}}
\toprule
Active$\leftrightarrow$Logical & \raggedright Op (Softmax) Arg (Softmax) & \raggedright Op (Softmax) Arg (Gumbel) & \raggedright Op (Gumbel) Arg (Softmax) & \raggedright Op (Gumbel) Arg (Gumbel) & \\
\midrule
-train & $1.00 \pm 0.00$ & $.086 \pm .172$ & $0.00 \pm 0.00$ & $0.00 \pm 0.00$ \\
-test IID & $1.00 \pm 0.00$ & $.088 \pm .176$ & $0.00 \pm 0.00$ & $0.00 \pm 0.00$ \\
-test OOD lexical & $1.00 \pm 0.00$ & $.094 \pm .188$ & $0.00 \pm 0.00$ & $0.00 \pm 0.00$ \\
-test OOD structural & $1.00 \pm 0.00$ & $.068 \pm .136$ & $0.00 \pm 0.00$ & $0.00 \pm 0.00$ \\
\bottomrule
\end{tabular}
\end{sc}
\end{center}
\vskip -0.1in
\caption{Accuracy on Active$\leftrightarrow$Logical across five random initializations for models which use varying combinations of softmax and Gumbel-Softax for operation and argument selection.}
\label{tab:gumbel}
\end{table*}


\subsection{Interpreting Inference as Programs} \label{sec:interpret}
\begin{figure}[]
\begin{center}
\centerline{\includegraphics[width=\columnwidth]{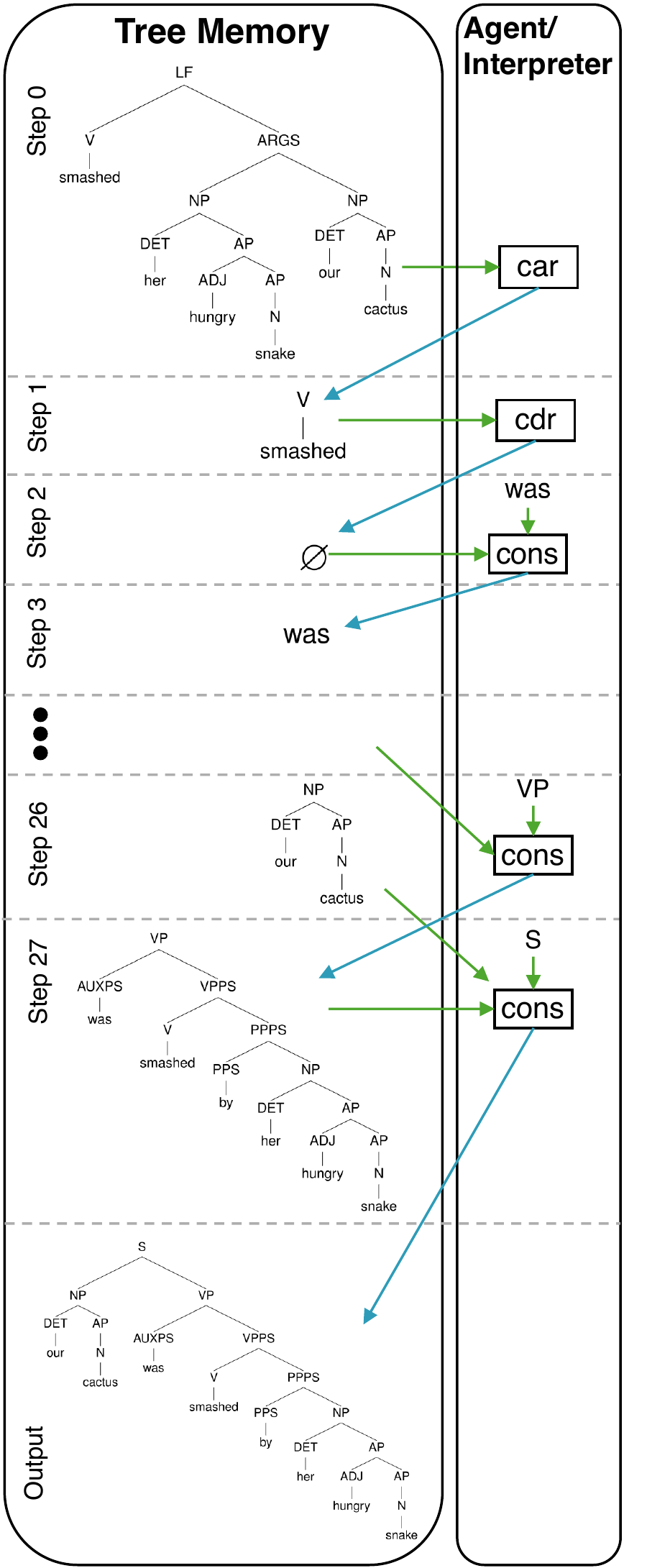}}
\caption{An interpretable transformation from logical form to passive. For readability, trees are shown here symbolically, but Tree Memory contains the vector embeddings (TPRs) of these trees. At each step, all of the items in memory from previous steps are available to the agent/interpreter. Reads are shown in green and writes in blue. The interpretation is discussed in Section \ref{sec:interpret}.}
\label{fig:interpret}
\end{center}
\vskip -0.2in
\end{figure}

The output of the Neural Tree Agent at each timestep can be interpreted as routing data and performing a predefined operation. At convergence, we find that the path from the input tree to the output tree is defined by interpretable one-hot softmax distributions. For the \textbf{CAR-CDR-SEQ} task, our model learns to act as a program interpreter and performs the correct sequence ($\sim$94\% accuracy) of operations on subsequent trees throughout computation. For the language tasks, we can trace the program execution to see how the input tree is transformed into the output tree. An example of our model's behavior over 28 steps on Logical$\rightarrow$Passive can be seen in Figure \ref{fig:interpret}. In particular, we were excited to find an emergent operation in our model's behavior. Transducing from Logical$\rightarrow$Passive not only requires rearranging nodes but also inserting new words into the tree, ``was" and ``by". At first glance, \car, \cdr, and \cons\ do not appear to support adding a new node to memory. The model learns that taking \cdr\ of a tree with only a single child returns an empty tree (Step 2 in Figure \ref{fig:interpret}); the empty tree can then be used as the inputs to \cons\ in order to write a new word as the root node with no children on the left or right (Step 3). The programmatic nature of our network at convergence --- the fact that the weighting coefficients $\vec{w}, \vec{a}$ become 1-hot --- makes it trivial to discover how an undefined operation emerged during training.

\section{Conclusions, Limitations, and Future Work}

We introduce \blackboard, an architecture for leveraging differentiable tree operations and an external memory to achieve compositional generalization. Our trees are embedded in vector space as TPRs which allow us to perform symbolic operations as differentiable linear transformations. \blackboard\ achieves 100\% out-of-distribution generalization for both lexical and structural distributional shifts across a variety of synthetic tree-to-tree tasks and is highly interpretable.

The major limitation of \blackboard\ is that the input and output must be structured as trees. Future work will focus on allowing \blackboard\ to take in unstructured data, generate a tree, and then perform the operations we described. This will allow \blackboard\ to be evaluated on a larger variety of datasets. Our hope is that \blackboard\ will be able to scale to language modeling and other large pretraining tasks. Our model is also restricted to tree transformations where the input and output languages share the same vocabulary. While these sorts of transformations are common in Computer Science and Natural Language Processing, many tasks involve translating between vocabularies, and future work will investigate ways to translate between different vocabularies.

Another hurdle for \blackboard\ is the size of trees that it can handle. The encoding scheme we presented here requires the depth of possible trees to be determined beforehand so that the appropriate number of roles can be initialized. Additionally, while the TPR dimension grows linearly with the number of nodes in a tree, the number of nodes in a tree grows exponentially with depth. The majority of \blackboard\ parameters reside in $W_{shrink}$, the linear transformation from TPR space to model space. This can cause memory issues when representing deep trees. We leave methods for extending to an unbounded depth and lossy compression of the role space to future work.

Finally, while \blackboard\ reduces the operation space to individual transformations, it may be feasible for a model to learn more expressive functions which combine multiple \car, \cdr, and \cons\ operations in a single step. Future work will investigate other tree functions, such as Tree Adjoining, as well as other data structures. The sequences of all possible operations define an infinite set of functions which are linear transformations, and we hope that our work will inspire further research into this space.

In conclusion, we believe that \blackboard\ represents a promising direction for leveraging differentiable tree operations and external memory to achieve compositional generalization. Our model is interpretable, systematic, and has potential for scaling to larger datasets and different data structures. We hope that our work will inspire further research in this area and facilitate progress towards building more powerful and interpretable models for structured data.
\section{Impact Statement}
To our knowledge, the work presented here poses no societal harms beyond the scope of general AI research. As with all ML research, there is a risk that improvements in ML  technologies could be used for harmful purposes. We hope that the interpretability offered by our method can contribute to an understanding of neural network models to increase controllability as well as improve and verify fairness.

\section*{Acknowledgements}
We are grateful to the Johns Hopkins Neurocompositional Computation group, the Microsoft Research Redmond Deep Learning Group, and the anonymous reviewers for helpful comments. We are also grateful for the feedback provided by Colin Wilson, Ricky Loynd and Steven Piantadosi. Soulos was partly supported by the Cognitive Science Department at Johns Hopkins. Any errors remain our own.




\bibliography{tpbb}
\bibliographystyle{icml2023}

\newpage
\appendix
\onecolumn
\section{Model Hyperparameter Selection}
\raggedbottom

For all of the models we evaluated, the HP searching and training was done in 3 steps:
\begin{enumerate}
   \item An optional exploratory random search over a wide range of HP values (using the Active$\leftrightarrow$Logical task) 
   \item A grid search (repeat factor=3) over the most promising HP values from step 1 (using the Active$\leftrightarrow$Logical task) 
   \item Training on the target tasks (repeat factor=5) 
\end{enumerate}

All of our models were trained on 1x V100 (16GB) virtual machines.  

\subsection{\blackboard} \label{sec:blackboard-training}
For the \blackboard\ models, we ran a 3x hyperparameter grid search over the following ranges. The best performing hyperparameter values are marked in bold.

\begin{tabular}{ll}
    Computation Steps: & [X+2, \textbf{(X+2)*2}] where X is the minimum number of steps required to complete a task \\  
    weight\_decay: & [\textbf{.1}, .01] \\
    Transformer model dimension: & [32, \textbf{64}] \\
    Adam $\beta_2$: & [.98, \textbf{.95}] \\
    Transformer dropout: & [0, \textbf{.1}] \\
\end{tabular}

The following hyperparameters were set for all models

\begin{tabular}{ll}
    lr\_warmup: & [10000] \\
     lr\_decay: & [cosine] \\
     training steps: & [20000] \\
     Transformer encoder layers per computation step: & [1] \\
     Transformer \# of heads: & [4] \\
     Batch size: & [16] \\
     d\_symbol: & \# symbols in the dataset \\
     d\_role: & $2^{D+1}-1$ where D is the max depth in the dataset \\
     Transformer non-linearity: & gelu \\
     Optimizer: & Adam \\
     Adam $\beta_1$: & .9 \\
     Gradient clipping: & 1 \\
     Transformer hidden dimension: & 4x Transformer model dimension \\
\end{tabular}

Notes:

\begin{itemize}
\item For the Passive$\leftrightarrow$Logical task, a batch size of 8 was used to reduce memory requirements. 
\item Training runs that didn't achieve 90\% training accuracy were excluded from evaluation 
\end{itemize}

\pagebreak
\subsection{Baselines}

We search over model and training hyperparameters and choose the combination that has the highest (and in the case of ties, quickest to train) mean validation accuracy on Active \& Passive$\rightarrow$Logical. The best hyperparameter setting for each model was then used to train that model on all four of our tasks.

\subsubsection{Transformer}

The Transformer 1x exploratory random search operated on the following HP values: 

\begin{tabular}{ll}
lr: & [.0001, .00005] \\
lr\_warmup: & [0, 1000, 3000, 6000, 9000] \\
lr\_decay: & [none, linear, factor, noam] \\
lr\_decay\_factor: & [.9, .95, .99] \\
lr\_patience: & [0, 5000] \\
stop\_patience: & [0] \\
weight\_decay: & [0, .001, .01] \\
hidden: & [64, 96, 128, 256, 512, 768, 1024] \\
n\_encoder\_layers: & [1, 2, 3, 4, 5, 6, 7, 8] \\
n\_decoder\_layers: & [1, 2, 3, 4, 5, 6, 7, 8] \\
dropout: & [0, .1, .2, .3, .4] \\
filter: & [256, 512, 768, 1024, 2048, 3096, 4096] \\
n\_heads: & [1, 2, 4, 8, 16] \\     
\end{tabular}

\vskip 0.15in
The Transformer 3x grid search operated on the following HP values: 

\begin{tabular}{ll}
    hidden: & [768, 1024] \\
    n\_encoder\_layers: & [1, 4] \\
    n\_decoder\_layers: & [3, 4] \\
    dropout: & [0] \\
    filter: & [768, 1024] \\
    n\_heads: & [2, 4] \\   
\end{tabular}
\vskip 0.15in

The Transformer 5x training on the target tasks was done with these final HP values:

\begin{tabular}{ll}
    n\_steps: & 30\_000 \\  
    log\_every: & 100 \\
    eval\_every: & 1000 \\
    batch\_size\_per\_gpu: & 256 \\
    max\_tokens\_per\_gpu: & 20\_000   \\
    lr: & .0001 \\
    lr\_warmup: & 1000 \\
    lr\_decay: & linear \\
    lr\_decay\_factor: & .95 \\
    lr\_patience: & 5000 \\
    stop\_patience: & 0 \\
    optimizer: & adam \\    
    weight\_decay: & 0 \\
    max\_abs\_grad\_norm: & 1 \\
    grad\_accum\_steps: & 1 \\
    greedy\_must\_match\_tf: & 0 \\
    early\_stop\_perfect\_eval: & 0 \\
    hidden: & 1024 \\
    n\_encoder\_layers: & 1 \\
    n\_decoder\_layers: & 3 \\
    dropout: & 0 \\
    filter: & 1024 \\
    n\_heads: & 2 \\     
\end{tabular}

\pagebreak
\subsubsection{LSTM}

The LSTM 1x exploratory random search operated on the following HP values:

\begin{tabular}{ll}

    weight\_decay: & [0, .001, .01] \\
    lr: & [.0001, .00005] \\
    lr\_warmup: & [0, 1000, 3000, 6000, 9000] \\
    lr\_decay: & [none, linear, factor, noam] \\
    lr\_decay\_factor: & [.9, .95, .99] \\
    lr\_patience: & [0, 5000] \\
    hidden: & [64, 96, 128, 256, 512, 768, 1024] \\
    n\_encoder\_layers: & [1, 2, 3, 4, 5, 6] \\
    n\_decoder\_layers: & [1, 2, 3, 4, 5, 6] \\
    dropout: & [0, .05, .1, .15, .2] \\
    bidir: & [0, 1] \\
    use\_attn: & [0, 1] \\
    rnn\_fold: & [min, max, mean, sum, hadamard] \\
    attn\_inputs: & [0, 1] \\  
\end{tabular}
\vskip 0.15in

The LSTM 3x grid search operated on the following HP values:

\begin{tabular}{ll}
    lr\_decay: & [linear, noam] \\
    hidden: & [512, 768, 1024] \\
    n\_encoder\_layers: & [1, 6] \\
    n\_decoder\_layers: & [1, 2, 3] \\
    attn\_inputs: & [0, 1] \\  
\end{tabular}
\vskip 0.15in

The LSTM 5x training on the target tasks was done with these final HP values:

\begin{tabular}{ll}
    n\_steps: & 30\_000 \\  
    log\_every: & 200 \\
    eval\_every: & 1000 \\
    stop\_patience: & 0 \\
    optimizer: & adam \\    
    max\_abs\_grad\_norm: & 1 \\
    grad\_accum\_steps: & 1 \\
    greedy\_must\_match\_tf: & 0 \\
    early\_stop\_perfect\_eval: & 0 \\
    batch\_size\_per\_gpu: & 256 \\
    max\_tokens\_per\_gpu: & 20\_000  \\  
    weight\_decay: & 0 \\
    lr: & .0001 \\
    lr\_warmup: & 1000 \\
    lr\_decay: & noam \\
    lr\_decay\_factor: & .95 \\
    lr\_patience: & 5000 \\
    hidden: & 512 \\
    n\_encoder\_layers: & 6 \\
    n\_decoder\_layers: & 1 \\
    dropout: & .1 \\
    bidir: & 0 \\
    use\_attn: & 1 \\
    rnn\_fold: & max \\
    attn\_inputs: & 1 \\  
\end{tabular}

\subsubsection{Tree2Tree}

The Tree2Tree 1x exploratory random search operated on the following HP values: 

\begin{tabular}{ll}
lr: & [.01, .005, .001, .0005] \\
lr\_decay\_factor: & [.8, .9, .95] \\
max\_abs\_grad\_norm: & [1, 5] \\
hidden: & [64, 128, 256, 512, 768] \\
dropout: & [0, .1, .2, .4, .5, .6] \\
lr\_decay: & [none, linear, factor, patience, noam] \\
\end{tabular}

\vskip 0.15in
The Tree2Tree 3x grid search operated on the following HP values: 

\begin{tabular}{ll}
    dropout: & [0, .6] \\
    hidden: & [512, 768] \\
\end{tabular}
\vskip 0.15in

The Tree2Tree 5x training on the target tasks was done with these final HP values:

\begin{tabular}{ll}
    n\_steps: & [10\_000]  \\
    stop\_patience: & [0] \\
    early\_stop\_perfect\_eval: & [0] \\
    lr: & [.0005] \\
    lr\_decay: & [patience] \\
    lr\_warmup: & [1000] \\
    lr\_patience: & [500] \\
    lr\_decay\_factor: & [.95] \\
    batch\_size\_per\_gpu: & [256] \\
    max\_tokens\_per\_gpu: &  null    \\
    optimizer: & [adam]     \\
    weight\_decay: & [0] \\
    max\_abs\_grad\_norm: & [1] \\
    grad\_accum\_steps: & [1] \\
    n\_encoder\_layers: & [1] \\
    n\_decoder\_layers: & [1] \\
    dropout: & [0] \\
    hidden: & [512] \\
\end{tabular}

\pagebreak

\subsubsection{TreeTransformer}

The TreeTransformer 1x exploratory random search operated on the following HP values: 

\begin{tabular}{ll}
    dropout\_rate: & [0, .05, .1, .2] \\
    batch\_size: & [64, 128, 256] \\
    learning\_rate: & [.0001, .0005, .001, .00001] \\
    optimizer: & [adam, sgd, momentum, adagrad, adadelta, rmsprop] \\
    max\_gradient\_norm: & [0.0, .05, .1] \\
    momentum: & [0.0, .1, .5, .9] \\
    d\_model: & [128, 256] \\
    d\_ff: & [128, 256, 512, 1024] \\
    encoder\_depth: & [1, 2, 3, 4] \\
    decoder\_depth: & [2, 4, 6, 8] \\
    
\end{tabular}

\vskip 0.15in
The TreeTransformer 3x grid search operated on the following HP values: 

\begin{tabular}{ll}
    optimizer: & [adagrad, adadelta] \\
    max\_gradient\_norm: & [.05, .1] \\
    momentum: & [0.0, .5] \\
    d\_model: & [256] \\
    d\_ff: & [256] \\
    encoder\_depth: & [1, 2] \\
    decoder\_depth: & [2, 4] \\
\end{tabular}
\vskip 0.15in

The TreeTransformer 5x training on the target tasks was done with these final HP values:

\begin{tabular}{ll}
    train\_batches: & [30\_000]   \\
    max\_eval\_steps: & [2000] \\

    dropout\_rate: & [0] \\
    batch\_size: & [256] \\
    learning\_rate: & [.0001] \\
    optimizer: & [adagrad] \\
    max\_gradient\_norm: & [.05] \\
    momentum: & [.5] \\
    num\_heads: & [2] \\
    d\_model: & [256] \\
    d\_ff: & [256] \\
    encoder\_depth: & [1] \\
    decoder\_depth: & [2] \\
\end{tabular}

\vskip 0.15in

\pagebreak
\section{Basic Sentence Transforms}

\subsection{Dataset Construction} \label{sec:dataset-construction}
The Basic Sentence Transforms vocabulary size and tree depth are available below. The lexical splits are constructed by using 1 set of adjectives for the Train, Dev, Test IID, and OOD Structural splits, and a disjoint set for the OOD Lexical split. The structural splits are constructed by using 0-2 nested adjectives distributed randomly to the two noun phrases for train, dev, and OOD Lexical splits, and 3-4 nested adjectives to the two noun phrases for the OOD Structural split. Adjective phrases are nested within each other within a noun phrase, so each additional adjective increases the overall tree depth by 1.

\textbf{Vocabulary Size}
\begin{table}[H]
\begin{center}
\begin{sc}
\begin{tabular}{lccc}
\toprule
Dataset & Train/Dev/Test & Test OOD Lexical & Test OOD Structural \\
\midrule
CAR\_CDR\_SEQ & 142 & 153 & 142 \\
Active$\leftrightarrow$Logical & 101 & 112 & 101 \\
Passive$\leftrightarrow$Logical & 107 & 118 & 107 \\
Active \& Passive$\rightarrow$Logical & 105 & 116 & 105 \\
\bottomrule
\end{tabular}
\end{sc}
\end{center}
\vskip -0.1in
\end{table}

\textbf{Tree Depth}
\begin{table}[H]
\begin{center}
\begin{sc}
\begin{tabular}{lccc}
\toprule
Dataset & Train/Dev/Test & Test OOD Lexical & Test OOD Structural \\
\midrule
CAR\_CDR\_SEQ & 10 & 10 & 12 \\
Active$\leftrightarrow$Logical & 8 & 8 & 10 \\
Passive$\leftrightarrow$Logical & 10 & 10 & 12 \\
Active \& Passive$\rightarrow$Logical & 10 & 10 & 12 \\
\bottomrule
\end{tabular}
\end{sc}
\end{center}
\vskip -0.1in
\end{table}

\subsection{Dataset Samples} \label{sec:dataset-samples}
This appendix contains samples of the 4 tasks that we used from the Basic Sentence Transforms Dataset.

\subsubsection{\textbf{CAR-CDR-SEQ} Samples}

\underline{Source Tree:}

\hspace{.15in} ( CDDDDR ( NP ( DET the ) ( AP ( N goat ) ) ) ( VP ( AUXPS was ) ( VPPS ( V bought ) ( PPPS ( PPS by ) ( NP ( DET the ) ( AP ( ADJ round ) ( AP ( N rose ) ) ) ) ) ) ) )	

\begin{figure}[H]
\vskip 0.2in
\begin{center}
\centerline{\includegraphics[width=.4\columnwidth]{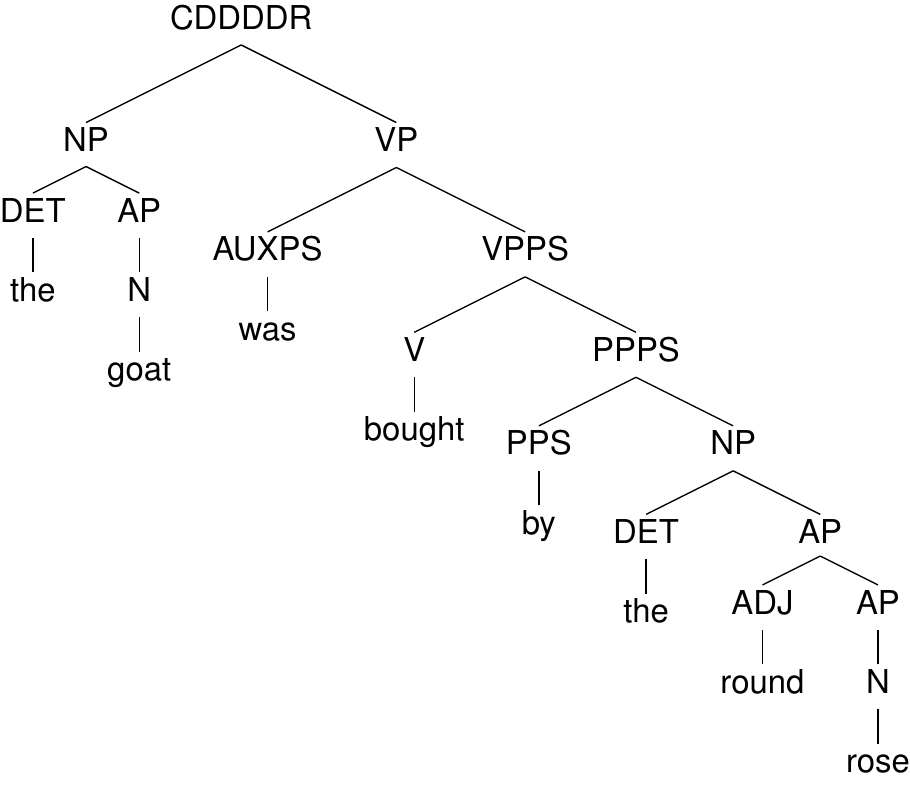}}
\end{center}
\vskip -0.2in
\end{figure}

\underline{Target (Gold) Tree:}

\hspace{.15in}( NP ( DET ( the ) ) ( AP ( ADJ ( round ) ) ( AP ( N ( rose ) ) ) ) )

\begin{figure}[H]
\vskip 0.2in
\begin{center}
\centerline{\includegraphics[width=.2\columnwidth]{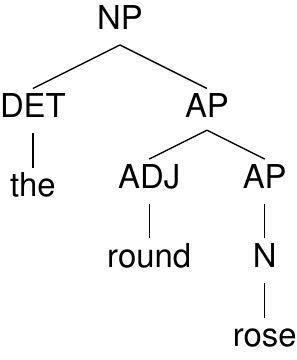}}
\end{center}
\vskip -0.2in
\end{figure}

\subsubsection{\textbf{Active$\leftrightarrow$Logical} Samples}

\underline{Source Tree:}

\hspace{.15in} ( S ( NP ( DET some ) ( AP ( N crocodile ) ) ) ( VP ( V washed ) ( NP ( DET our ) ( AP ( ADJ happy ) ( AP ( ADJ thin ) ( AP ( N donkey ) ) ) ) ) ) )	

\begin{figure}[H]
\vskip 0.2in
\begin{center}
\centerline{\includegraphics[width=.4\columnwidth]{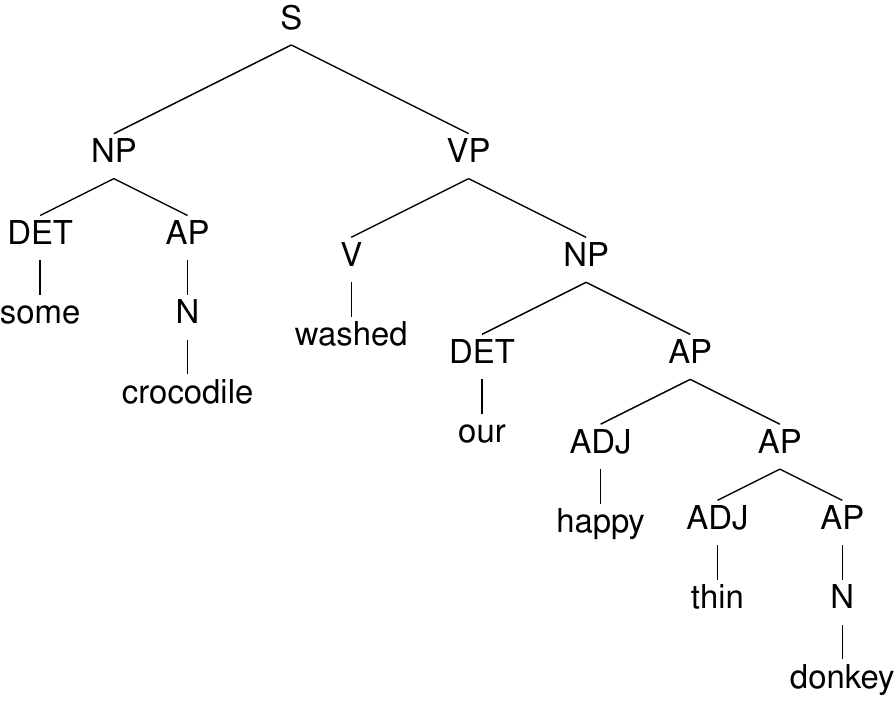}}
\end{center}
\vskip -0.2in
\end{figure}

\underline{Target (Gold) Tree:}

\hspace{.15in} ( LF ( V washed ) ( ARGS ( NP ( DET some ) ( AP ( N crocodile ) ) ) ( NP ( DET our ) ( AP ( ADJ happy ) ( AP ( ADJ thin ) ( AP ( N donkey ) ) ) ) ) ) )	

\begin{figure}[H]
\vskip 0.2in
\begin{center}
\centerline{\includegraphics[width=.4\columnwidth]{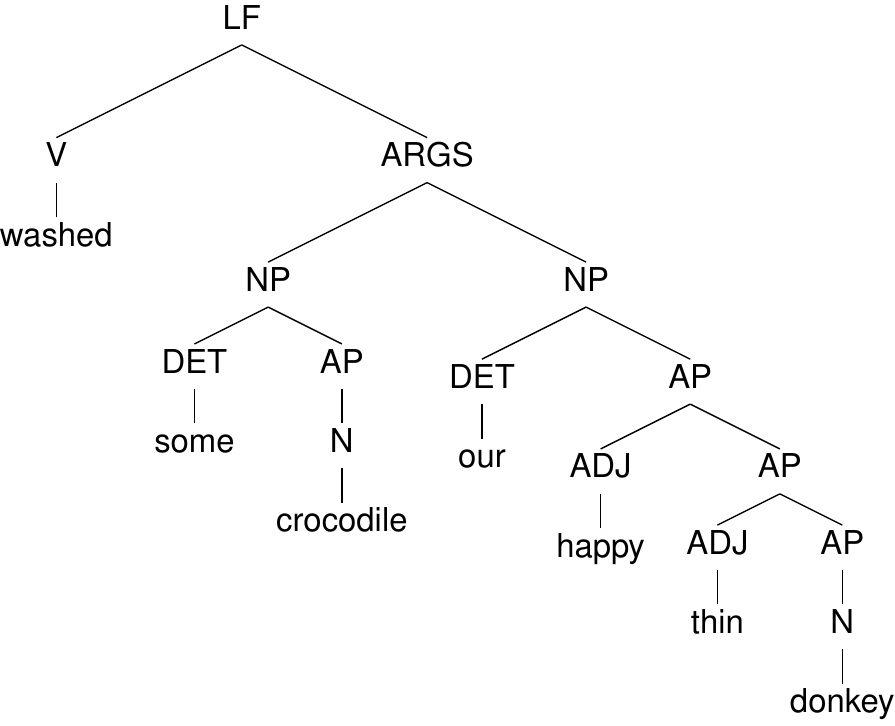}}
\end{center}
\vskip -0.2in
\end{figure}

\subsubsection{\textbf{Passive$\leftrightarrow$Logical} Samples}

\underline{Source Tree:}
\hspace{.15in} ( S ( NP ( DET his ) ( AP ( N tree ) ) ) ( VP ( AUXPS was ) ( VPPS ( V touched ) ( PPPS ( PPS by ) ( NP ( DET one ) ( AP ( ADJ polka-dotted ) ( AP ( N crocodile ) ) ) ) ) ) ) )	

\begin{figure}[H]
\vskip 0.2in
\begin{center}
\centerline{\includegraphics[width=.4\columnwidth]{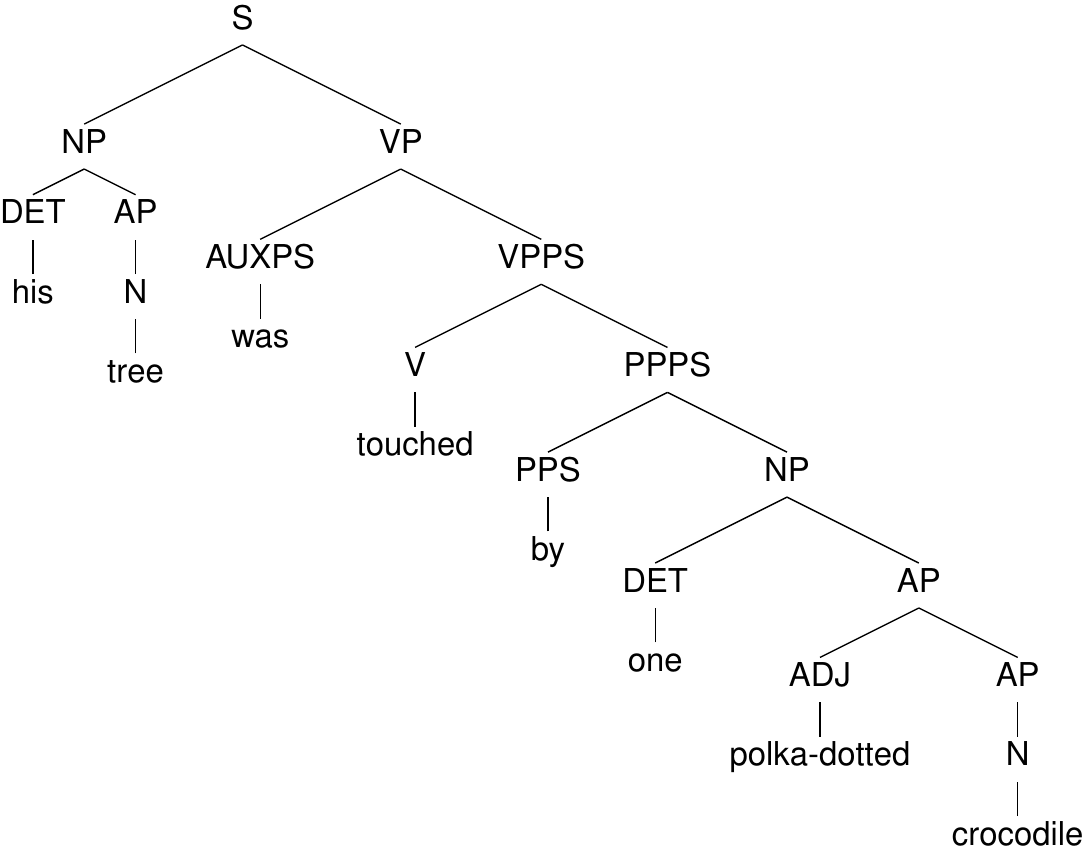}}
\end{center}
\vskip -0.2in
\end{figure}

\underline{Target (Gold) Tree:}
\hspace{.15in} ( LF ( V touched ) ( ARGS ( NP ( DET one ) ( AP ( ADJ polka-dotted ) ( AP ( N crocodile ) ) ) ) ( NP ( DET his ) ( AP ( N tree ) ) ) ) )

\begin{figure}[H]
\vskip 0.2in
\begin{center}
\centerline{\includegraphics[width=.4\columnwidth]{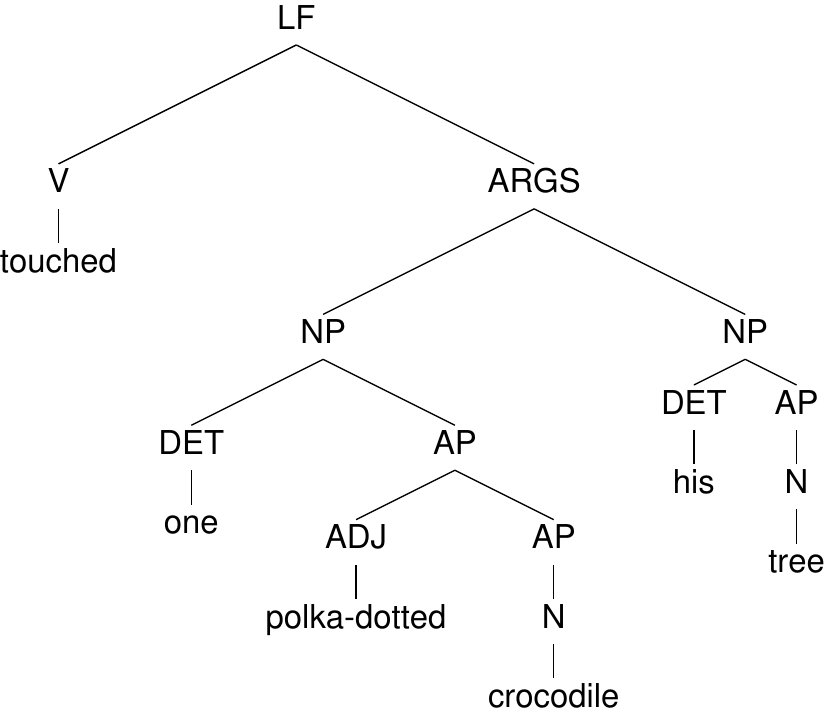}}
\end{center}
\vskip -0.2in
\end{figure}

\subsubsection{\textbf{Active \& Passive$\rightarrow$Logical} Samples}

\underline{Source Tree:}
\hspace{.15in} ( S ( NP ( DET a ) ( AP ( N fox ) ) ) ( VP ( AUXPS was ) ( VPPS ( V kissed ) ( PPPS ( PPS by ) ( NP ( DET my ) ( AP ( ADJ blue ) ( AP ( N giraffe ) ) ) ) ) ) ) )

\begin{figure}[H]
\vskip 0.2in
\begin{center}
\centerline{\includegraphics[width=.4\columnwidth]{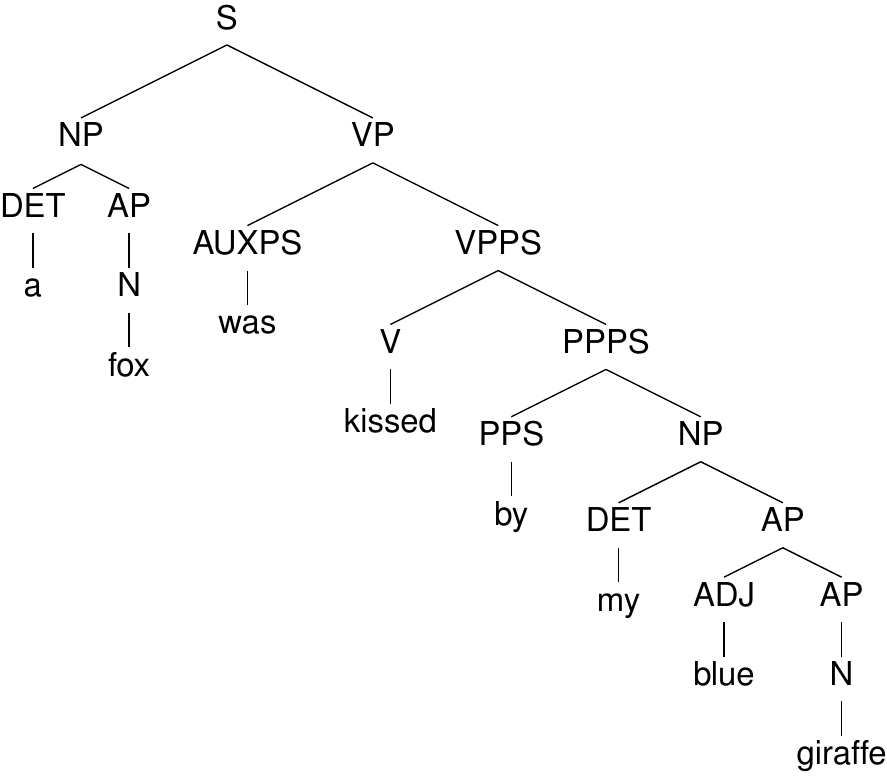}}
\end{center}
\vskip -0.2in
\end{figure}

\underline{Target (Gold) Tree:}
\hspace{.15in} ( LF ( V kissed ) ( ARGS ( NP ( DET my ) ( AP ( ADJ blue ) ( AP ( N giraffe ) ) ) ) ( NP ( DET a ) ( AP ( N fox ) ) ) ) )

\begin{figure}[H]
\vskip 0.2in
\begin{center}
\centerline{\includegraphics[width=.4\columnwidth]{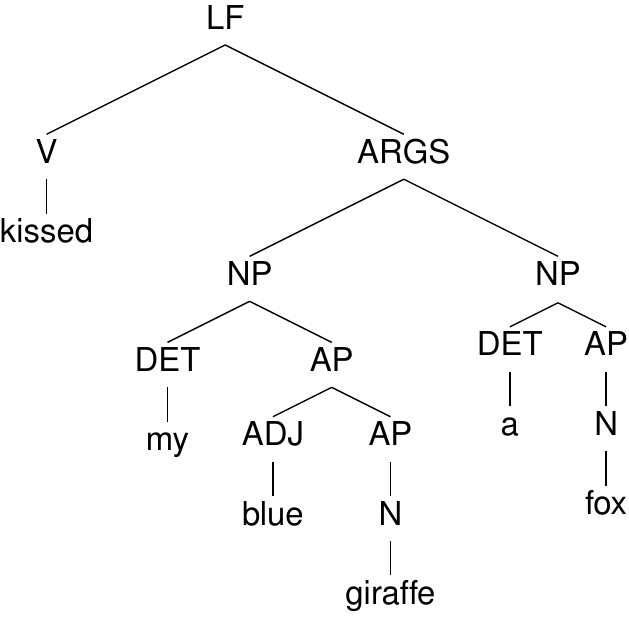}}
\end{center}
\vskip -0.2in
\end{figure}

\end{document}